%% file: main.tex
\documentclass[runningheads]{llncs}
\usepackage{graphicx}
\usepackage{booktabs}
\usepackage[utf8]{inputenc}
\usepackage[T1]{fontenc}
\usepackage{bm}
\usepackage{amsmath,amsfonts}
\usepackage{lineno}%
\usepackage{makecell}
\usepackage{scrextend}
\usepackage{float}
\usepackage{hyperref}

\newcommand{\bi}{\begin{itemize}}
\newcommand{\ei}{\end{itemize}}
\newcommand{\ba}{\begin{array}}
\newcommand{\ea}{\end{array}}

\usepackage{xspace}

\newcommand{\bmx}[0]{\begin{bmatrix}}
\newcommand{\emx}[0]{\end{bmatrix}}

\newif\ifarxiv
\arxivtrue

\begin{document}
\title{Evaluating Movement Initiation Timing in Ultimate Frisbee via Temporal Counterfactuals}
%
\titlerunning{Movement Initiation Timing in Ultimate Frisbee}
%

\ifarxiv
\author{Shunsuke Iwashita\inst{1} \and Ning Ding\inst{2} \and Keisuke Fujii\inst{1,3}}
%
\institute{Graduate School of Informatics, Nagoya University, Nagoya, Japan. \and
Graduate School of Engineering, Nagoya Institute of Technology, Nagoya, Japan. \and
Center for Advanced Intelligence Project, RIKEN, Osaka, Japan.
\email{iwashita.shunsuke@g.sp.m.is.nagoya-u.ac.jp,
fujii@i.nagoya-u.ac.jp}
}
\authorrunning{S. Iwashita et al.}
\else
\author{Anonymous}
\institute{}
\vspace{-11pt}

\fi
\maketitle              
\begin{abstract}
\vspace{-15pt}
Ultimate is a sport where points are scored by passing a disc and catching it in the opposing team's end zone. In Ultimate, the player holding the disc cannot move, making field dynamics primarily driven by other players' movements. However, current literature in team sports has ignored quantitative evaluations of when players initiate such unlabeled movements in game situations. In this paper, we propose a quantitative evaluation method for movement initiation timing in Ultimate Frisbee. First, game footage was recorded using a drone camera, and players' positional data was obtained, which will be published as UltimateTrack dataset. Next, players' movement initiations were detected, and temporal counterfactual scenarios were generated by shifting the timing of movements using rule-based approaches. These scenarios were analyzed using a space evaluation metric based on soccer's pitch control reflecting the unique rules of Ultimate. By comparing the spatial evaluation values across scenarios, the difference between actual play and the most favorable counterfactual scenario was used to quantitatively assess the impact of movement timing. 
We validated our method and show that sequences in which the disc was actually thrown to the receiver received higher evaluation scores than the sequences without a throw.
In practical verifications, the higher-skill group displays a broader distribution of time offsets from the model’s optimal initiation point.
These findings demonstrate that the proposed metric provides an objective means of assessing movement initiation timing, which has been difficult to quantify in unlabeled team sport plays.
\ifarxiv
\footnote{The code is available at \url{https://github.com/shunsuke-iwashita/VTCS}}
\else
\fi

\vspace{-6pt}
\end{abstract}
%
%
%


\vspace{-20pt}
\section{Introduction}
\vspace{-8pt}
\label{sec:introduction}
Ultimate Frisbee (hereafter ``Ultimate'') is a field sport played by two teams of seven athletes on a rectangular pitch that comprises a central zone flanked by two end zones. A point is scored when a player catches the flying disc in the opponent’s end zone. Since the thrower must establish a pivot and cannot run with the disc, territorial gain arises from passes directed into space created by the movements of off-ball (i.e., without the disc) teammates. 
In contrast with invasion games such as football, basketball, and American football, where extensive event and tracking data have provided a large analytics literature \cite{fujii2025machine}, empirical work on Ultimate has still been scarce. 
One of the key obstacles is the lack of publicly available player-tracking datasets, leaving many tactical questions unexplored.

Pioneering work of Ultimate tactics is Weiss and Childers's paper \cite{weiss2013maps}, which introduced location-based completion and scoring probability maps derived from location data to quantify the tactical value of each field region. This work was extended to the player contribution metric in \cite{weiss2014spatial}, but the tracking data was not published. 
Lam et al. \cite{lam2021state} modelled passes and turnovers as zone-to-zone state transitions, and Eberhard et al. \cite{eberhard2025machine} combined four seasons of professional tracking to build completion-probability and field-value models that jointly capture throw difficulty and positional value.
Most previous work in Ultimate has used only discrete event data, including the time and coordinates of each pass, catch, or turnover, without continuous xy trajectories for all players, because acquiring full-field tracking remains costly. 
Even in team sports that routinely collect tracking data, such as soccer and basketball, evaluating the optimal timing of unlabeled movement initiations poses two challenges. First, the onset of a run or cut must be detected in dense, overlapping trajectories (see also a specific play case in basketball \cite{Fujii16}). Second, such a moment must be linked to a motion model and a value function that can be trained with limited samples. Recently, the only study that has combined space valuation in Ultimate exists \cite{iwashita2024space}; their 3-vs-3 game tracking data modified an existing spatial metric \cite{OBSO} but did not address the timing problem.

In this paper, we propose VTCS (Valuing Timing by Counterfactual Scenarios), a framework that quantifies how much earlier or later a receiver should have initiated a movement. VTCS first detects each initiation frame for each player from player-tracking data, then generates a family of counterfactual plays by systematically shifting that frame forward and backward with a rule-based motion model. For every real and counterfactual sequence, it computes a frame-wise field value inspired by a pitch control model \cite{OBSO} but adapted to Ultimate’s mechanics. Then the timing score is defined as the gap between the realized trajectory and the best counterfactual. We model and evaluate VTCS on UltimateTrack, a drone-captured, frame-level dataset of 64 possessions of practice matches. VTCS is therefore the first approach to provide an objective, data-driven estimate of optimal movement-initiation timing in invasion sports where such unlabeled decisions were previously impossible to measure.

The main contributions of our paper are as follows. 
(1) We introduce the VTCS framework together with the open UltimateTrack dataset, enabling fine-grained timing evaluation in Ultimate for the first time. 
(2) We propose a rule-based counterfactual motion model, a sport-specific field-value formulation and a timing-effect metric defined as the gap between actual and optimal scenarios.
(3) Experiments show that thrown sequences score higher than non-throw sequences and that elite players exhibit a wider, reasonable spread around the model-optimal initiation time.
In the following sections, we describe 
our methods in Section \ref{sec:method} and the experimental results in Section \ref{sec:result}, and conclude this paper in Section \ref{sec:conclusion}.
Ultimate frisbee rules and related work are in Appendices \ref{app:ultimate} and \ref{app:related}.

\vspace{-12pt}
\section{Methods}
\vspace{-8pt}
\label{sec:method}
    In this study, we propose VTCS (Valuing Timing by Counterfactual Scenarios), a framework for quantitatively evaluating the optimality of a player’s initiation timing. This section describes each component of the VTCS framework. We begin by introducing the dataset and the motion model used to generate counterfactual scenarios. We then define the space evaluation metric. Finally, we present the timing evaluation metrics based on frame-level and scenario-level spatial values, which together measure how effective the actual initiation timing was.
    \vspace{-9pt}
    \subsection{UltimateTrack Datasets}
    \vspace{-5pt}
    \label{ssec:datasets}
        This study constructs and publishes a new tracking dataset called UltimateTrack, aimed at evaluating counterfactual scenarios and spatial control.

        The dataset was created based on aerial footage captured by a drone (DJI, Mavic 3) during a scrimmage conducted by the 
        \ifarxiv
        Nagoya 
        \fi
        University Flying Disc Team. 
        Player positions were manually tracked for every frame. The dataset is provided in its final processed form, incorporating unified representations of position, velocity, and acceleration, as described in the following section. It comprises 18,075 frames at 15 FPS and includes 64 possessions. The tracked objects consist of 15 entities: 7 offense players, 7 defense players, and 1 disc. All positional data are expressed in a normalized field coordinate system of 94  $\times$ 37 m. The data is formatted as a CSV file, where each row records the state of one object at one frame. The variables are listed in Appendix \ref{app:csv}.
        An example video is given \href{https://anonymous.4open.science/r/VTCS_appendix/UltimateTrack_sample.mp4}{here}.
        To the best of our knowledge, this is the first publicly available dataset in Ultimate Frisbee that provides continuous positional data for all players. 
        The preprocessing procedure is described in Appendix \ref{app:preprocess}.

        In this study, we target the receiver’s first decisive cut, defined as the moment a receiver accelerates for a potential pass. See Appendix \ref{app:detection} for details. In short, candidate initiations are detected automatically by simple kinematic rules that look for a sudden burst of acceleration aligned with the current velocity and preceded by a period in which the player has not possessed the disc. Subsequently, 455 sequences in total were visually reviewed to filter out movements toward crowded areas or those lacking a clear spatial objective. As a result, 310 sequences were retained for further analysis.
        
        While the present dataset was collected from a single team in a limited match-like setting, future work will include expanding data collection to a wider range of competitive environments and tactical contexts. We also plan to explore methods for streamlining annotation and scaling up data acquisition, potentially leveraging advances in computer vision. Although these extensions are not covered in this paper, we recognize them as important next steps for broadening the applicability and robustness of our approach.

    \vspace{-9pt}
    \subsection{Motion Model to Generate Counterfactual Scenarios}
    \vspace{-5pt}
    \label{ssec:counterfactual}
        To quantitatively evaluate the impact of a receiver’s initiation timing on the unfolding of gameplay, we construct \textbf{temporally counterfactual scenarios}. In these scenarios, only the initiation timing of the target player and their corresponding defender is altered, while all other conditions remain identical to those of the actual play. This framework enables the analysis of the causal effect that the timing of initiation has on spatial control. In designing our framework, we deliberately adopt a rule-based approach for generating counterfactual scenarios, rather than relying on fully data-driven \cite{teranishi2022evaluation,fujii2024decentralized,fujii2024estimating} or reinforcement learning-based models \cite{nakahara2023action,fujii2024adaptive,yeung2025openstarlab}. This choice allows explicit manipulation of the target player’s initiation timing while keeping all other variables fixed, making the causal effect both intuitive and interpretable. Furthermore, rule-based modeling is advantageous for providing actionable and comprehensible feedback to coaches and players, such as indicating ``a better outcome could have been achieved if initiation had occurred at this timing''. This interpretability is especially important under data constraints and in practical sports analysis.

        A series of counterfactual scenarios is constructed by altering the initiation timing of the detected player and considering the corresponding defender marking them. Specifically, based on the original initiation frame $t_0$, a set of temporally shifted position sequences $\{\mathbf{p}_i^{(\xi)}(t)\}$ is generated by applying a shift parameter $\xi \in [-15, 15]$, where $\xi < 0$ denotes an earlier and $\xi > 0$ denotes a delay initiation.
        \begin{figure}[htbp]
            \centering
            \begin{minipage}[b]{0.48\textwidth}
                \centering
                \includegraphics[width=\textwidth]{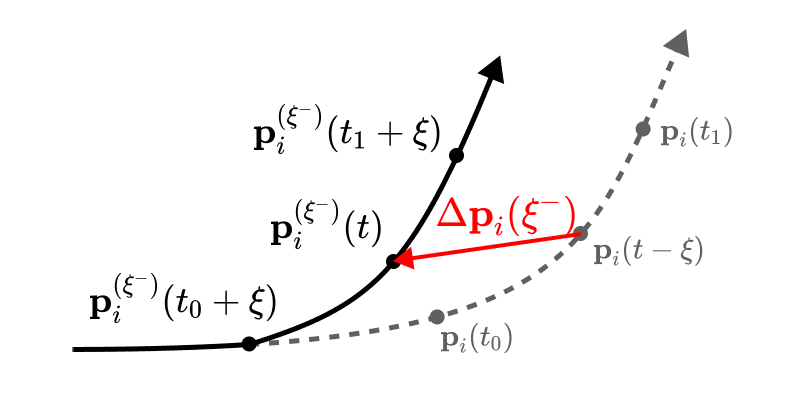}
                {\small (a) Initiation is brought forward ($\xi < 0$): The trajectory is shifted earlier and corrected by $\Delta \mathbf{p}_i(\xi^-)$}
            \end{minipage}
            \hfill
            \begin{minipage}[b]{0.48\textwidth}
                \centering
                \includegraphics[width=\textwidth]{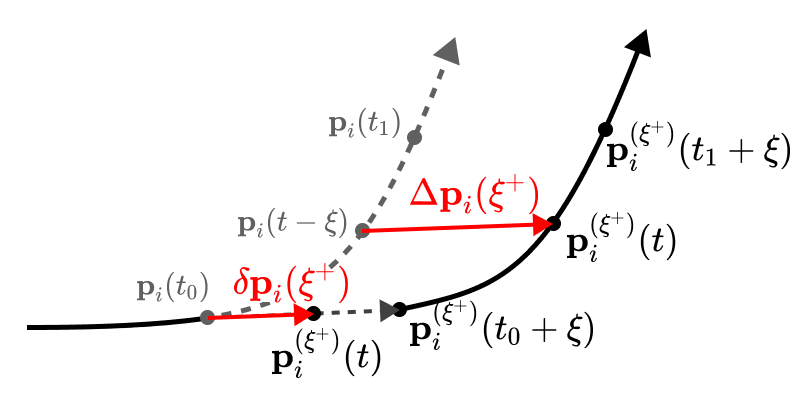}
                {\small (b) Initiation is delayed ($\xi > 0$): The gap is filled with $\delta \mathbf{p}_i(\xi^+)$ and corrected by $\Delta \mathbf{p}_i(\xi^+)$}
            \end{minipage}
            \vspace{-5pt}
            \caption{Visualization of temporally counterfactual scenarios (solid lines) by shifting the initiation timing of the target player (dashed line).}
            \label{fig:counterfactual_scenarios}
        \end{figure}

        First, when $\xi < 0$ (see Fig.~\ref{fig:counterfactual_scenarios} (a)), the player's original movement is replayed $|\xi|$ frames earlier, and the trajectory is translated to ensure continuity at the initiation point. The position sequence is defined as follows:
        \begin{equation}
            \mathbf{p}_i^{(\xi^-)}(t) =
            \begin{cases}
                \mathbf{p}_i(t) & \text{for } t \leq t_0 + \xi \\
                \mathbf{p}_i(t - \xi) + \Delta \mathbf{p}_i(\xi^-) & \text{for } t > t_0 + \xi
            \end{cases}
        \end{equation}
        To ensure continuity of the trajectory at the initiation point, a correction vector $\Delta \mathbf{p}_i(\xi^-) = \mathbf{p}_i(t_0 + \xi) - \mathbf{p}_i(t_0)$ is added after time $t_0 + \xi$. This adjustment ensures that the earlier-shifted trajectory connects smoothly with the original movement.

        On the other hand, when $\xi > 0$ (see Fig.~\ref{fig:counterfactual_scenarios} (b)), the player first undergoes a hypothetical linear motion $\delta \mathbf{p}_i(\xi^+)$ based on the average velocity $\bar{\mathbf{v}}_i$, and then follows the original trajectory delayed by $\xi$ frames. The corresponding position sequence is defined as:
        \begin{equation}
            \mathbf{p}_i^{(\xi^+)}(t) =
            \begin{cases}
                \mathbf{p}_i(t) & \text{for } t \leq t_0 \\
                \mathbf{p}_i(t_0) + \delta \mathbf{p}_i(\xi^+) & \text{for } t_0 < t \leq t_0 + \xi \\
                \mathbf{p}_i(t - \xi) + \Delta \mathbf{p}_i(\xi^+) & \text{for } t > t_0 + \xi
            \end{cases}
        \end{equation}
        The hypothetical movement term $\delta \mathbf{p}_i(\xi^+)$ represents a linear motion initiated from the original position $\mathbf{p}_i(t_0)$, proceeding at the average velocity $\bar{\mathbf{v}}_i$. This term is introduced to fill the temporal gap caused by the delayed initiation and to ensure continuity of motion in the counterfactual trajectory. The associated hypothetical movement and correction terms are defined as:
        \begin{equation}
            \delta \mathbf{p}_i(\xi^+) = \bar{\mathbf{v}}_i \cdot (t - t_0), \quad
            \Delta \mathbf{p}_i(\xi^+) = \bar{\mathbf{v}}_i \cdot \xi
        \end{equation}
        The average velocity vector over the one-second period (15 frames) prior to initiation is computed as:
            $ \bar{\mathbf{v}}_i = \frac{1}{15} \sum_{k=1}^{15} \mathbf{v}_i(t_0 - k)$.
        
        For all other players $j \neq i$, the position sequence remains identical to that of the original play:
            $ \mathbf{p}_j^{(\xi)}(t) = \mathbf{p}_j(t)$.
        This setup ensures that the only factor affecting spatial control in each counterfactual scenario is the initiation timing $\xi$ of the target player.
        The linear motion model provides a tractable baseline, but incorporating richer defensive dynamics is left for future research.

    \vspace{-9pt}
    \subsection{Space Evaluation Metric}
    \vspace{-5pt}
    \label{ssec:space_eval}
        To evaluate the impact of initiation timing on spatial advantage, we build upon the pitch control framework \cite{OBSO}, which defines the original Potential Pitch Control Field (PPCF).
        We adopt an Ultimate-specific extension of it, UPPCF \cite{iwashita2024space}, with a modified reaction time estimation derived from players’ velocity direction.
        Furthermore, we propose a weighted version, wUPPCF, which accounts for practical gameplay factors such as pass difficulty and defensive pressure.
        The definitions of PPCF and UPPCF are provided in Appendix \ref{app:space}.

        \vspace{-12pt}
        \subsubsection{Weighted Extension: wUPPCF.}
        \label{sssec:wuppcf}
            To further account for the difficulty of passing in Ultimate, we apply a distance-based weight $w_d$, which penalizes locations that are farther from the disc holder. Specifically, $w_d$ decreases as the distance from the disc increases, reflecting the decreased feasibility of successful long passes due to throwing limitations.
            This weighting encourages the evaluation to prioritize reachable and tactically viable spaces. The spatial distribution of $w_d$ is visualized in Appendix~\ref{app:distance_weight}, where warmer colors indicate higher feasibility.

            We also model the obstruction caused by a marker's blocking motion by defining a pair of virtual arms extending laterally from the defender and evaluating whether they intersect with the intended disc trajectory. The virtual arm length $r$ is defined as a function of the distance between the disc and the target:
            \begin{equation}
                r=1-\min \left(\frac{\|\mathbf{p}_t - \mathbf{p}_d\|}{30}, 1 \right)
            \end{equation}
            Here, $\mathbf{p}_d$ denotes the position of the disc, and $\mathbf{p}_t$ denotes the target location of the intended pass. If an intersection occurs between the virtual arms and the disc’s intended path, the obstruction weight by the marker's blocking motion $w_s$ is calculated based on the normalized distance from the defender to the intersection point. A shorter distance results in a stronger obstruction effect (lower weight), while a longer distance has less impact. If no intersection occurs, the weight is set to 1.

            Finally, the weighted spatial control metric for player $i$ is defined as:
            \begin{equation}
                \text{wUPPCF}_i = \text{UPPCF}_i \cdot w_d \cdot w_s
            \end{equation}
            wUPPCF is a practical spatial evaluation metric that extends PPCF for Ultimate by incorporating pass feasibility and defensive interference. It is computed individually for each player and is used in the following section to evaluate the effectiveness of initiation timing ($V_\text{timing}$).

    \vspace{-9pt}
    \subsection{Timing Evaluation Metric}
    \vspace{-5pt}
    \label{ssec:timing_metric}
        To evaluate the impact of initiation timing on spatial advantage, we propose a novel metric called VTCS. VTCS evaluates how effective a player's actual timing decision is by applying spatial control-based evaluation to a series of counterfactual scenarios constructed using the motion model (see Section~\ref{ssec:counterfactual}).
        VTCS is computed through the following three steps: frame-level evaluation metric $V_{\text{frame}}$, scenario-level evaluation metric $V_{\text{scenario}}$, and the final differential metric $V_\text{timing}$.

        \begin{figure}[t]
            \begin{minipage}[t]{0.48\textwidth}
                \centering
                \includegraphics[width=\linewidth]{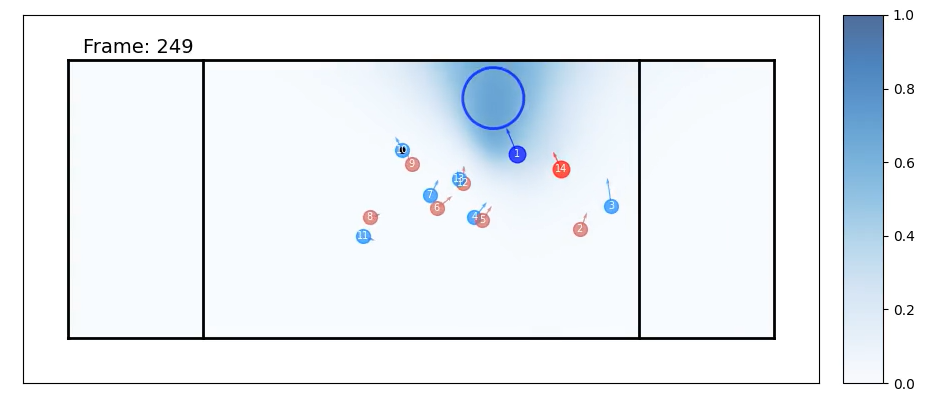}
                \vspace{-20pt}
                \caption{An example of 
                 frame-wise evaluation value $V_\text{frame}$, defined as the average wUPPCF within the predicted reachable area (blue circle) of the target receiver.}
                \label{fig:vframe}
            \end{minipage}
            \hfill
            \begin{minipage}[t]{0.48\textwidth}
                \centering
                \includegraphics[width=\linewidth]{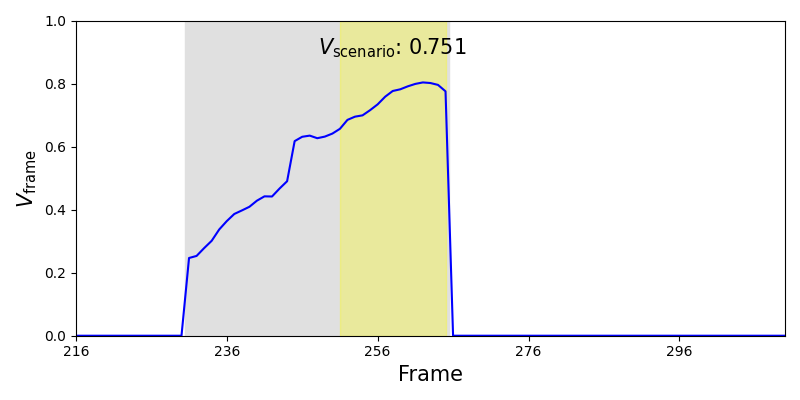}
                \vspace{-20pt}
                \caption{Temporal transition of $V_\text{frame}$. The scenario-wise evaluation value $V_\text{scenario}$ is defined as the maximum value of a 15-frame moving average applied to $V_\text{frame}$.}
                \label{fig:vscenario}
            \end{minipage}
        \end{figure}

        \vspace{-12pt}
        \subsubsection{$V_{\text{frame}}$.}
        \label{sssec:v_frame}
            The frame-wise value $V_{\text{frame}}(t)$ is defined as the average of wUPPCF (see Section~\ref{sssec:wuppcf}) within the area $\Omega(t)$, where the target player is likely to receive the disc:
            \begin{equation}
                V_{\text{frame}}(t) = \frac{1}{|\Omega(t)|} \sum_{\mathbf{r} \in \Omega(t)} \text{wUPPCF}_i(t, \mathbf{r})
            \end{equation}
            Here, $\Omega(t)$ denotes the region where the target player and the disc can arrive simultaneously. Specifically, we solve the following equations to find the intersection time $\tau$, based on the target player’s position and velocity $\mathbf{p}_i(t), \mathbf{v}_i(t)$, the disc’s position $\mathbf{p}_{\text{disc}}(t)$, and its known speed $v_{\text{disc}}$ with directional angle $\theta$:
            \begin{equation}
                \begin{cases}
                    p_{i,x}(t) + \tau \cdot v_{i,x}(t) = p_{\text{disc},x}(t) + \tau \cdot v_{\text{disc}} \cdot \cos\theta \\
                    p_{i,y}(t) + \tau \cdot v_{i,y}(t) = p_{\text{disc},y}(t) + \tau \cdot v_{\text{disc}} \cdot \sin\theta
                \end{cases}
            \end{equation}
            Using the solution $\tau$, we define the predicted position of the player as:
            \begin{equation}
                \mathbf{c}(t) = \mathbf{p}_i(t) + \tau \cdot \mathbf{v}_i(t)
            \end{equation}
            The region $\Omega(t)$ is then defined as a circle centered at $\mathbf{c}(t)$ with a radius of $\frac{1}{2}\|\mathbf{v}_i(t)\| \tau$, representing the area where the player is expected to be able to receive the disc. See Appendix~\ref{app:reachable_area} for an illustration.

        \vspace{-12pt}
        \subsubsection{$V_{\text{scenario}}$.}
        \label{sssec:v_scenario}
            For each counterfactual scenario with a timing shift $\xi$, the scenario-wise evaluation value $V_{\text{scenario}}^{(\xi)}$ is defined as the maximum of the moving average of $V_{\text{frame}}(t)$ over a 15-frame window:
            \begin{equation}
                V_{\text{scenario}}^{(\xi)} = \max_t \frac{1}{15} \sum_{k=1}^{15} V_{\text{frame}}(t + k)
            \end{equation}
            This allows us to identify the frame at which the target player gained the most spatial advantage, depending on the initiation timing.

        \vspace{-12pt}
        \subsubsection{$V_\text{timing}$.}
        \label{sssec:v_timing}
            $V_\text{timing}$ quantifies the effectiveness of the actual initiation timing as the difference between the scenario score of the actual case ($\xi = 0$) and the best counterfactual.
            \begin{equation}
            \text{$V_\text{timing}$} = V_{\text{scenario}}^{(0)} - \max_{\xi \in [-15, 15],\, \xi \neq 0} V_{\text{scenario}}^{(\xi)}
            \end{equation}
            Here, $V_{\text{scenario}}^{(0)}$ denotes the evaluation score of the actual scenario, while $\max V_{\text{scenario}}^{(\xi)}$ represents the highest score among all counterfactual scenarios.
            A large $V_\text{timing}$ value implies that the actual initiation timing led to significantly greater spatial advantage compared to any alternative timing, suggesting that the decision was highly effective. Conversely, a small $V_\text{timing}$ indicates that a better initiation timing may have existed, indicating room for improvement.

\vspace{-12pt}
\section{Results}
\vspace{-8pt}
\label{sec:result}
We conducted experiments using the UltimateTrack dataset introduced in Section~\ref{ssec:datasets} to evaluate the effectiveness of our proposed method.  
First, we validate the frame-wise evaluation metric $V_{\text{frame}}$, and then the timing evaluation metric $V_\text{timing}$. 
The use case of evaluating the initiation timing of a receiver and an optimal alternative is presented in Appendix \ref{app:application}, with a video demonstration available \href{https://anonymous.4open.science/r/VTCS_appendix/application_sample.mp4}{here}. 

    \vspace{-12pt}
    \subsection{Validation of frame-wise evaluation metric $V_{\text{frame}}$}
    \vspace{-5pt}
    \label{ssec:validation_vframe}
        Here we evaluate whether the proposed metric $V_{\text{frame}}$ provides a valid assessment that aligns with actual passing decisions. Specifically, we statistically analyze the relationship between $V_{\text{frame}}$ values and whether a pass was thrown to the detected player. Importantly, this comparison was conducted exclusively on the held-out test data from the same five-fold Group K-Fold cross-validation procedure, ensuring that no data leakage occurred.

        \vspace{-12pt}
        \subsubsection{Construction of Evaluation Data via Target Prediction Model.}
            In the context of temporally counterfactual scenarios, the timing of a player's initiation of movement is altered to generate multiple hypothetical trajectories. Consequently, whether a pass is thrown to the player in each scenario may vary, making it impossible to define a single ground truth for pass occurrence.
            To address this issue, we construct a classifier that estimates the probability that a detected player is the intended pass target in each frame, and use this prediction as a proxy for ground truth.
            Specifically, we train an XGBoost model using spatial and kinematic features derived from the scenario data. A 5-fold GroupKFold cross-validation is employed to assess generalization performance.
            The model achieved a mean RMSE of approximately 0.316 and a mean coefficient of determination ($R^2$) of approximately 0.163, indicating moderate predictive performance. Given the inherent uncertainty near the decision boundary, we exclude ambiguous cases and retain only those instances where the predicted probability was clearly high ($\geq$ 0.55) or low ($\leq$ 0.30). This filtering enables a more reliable comparison of $V_{\text{frame}}$ values in relation to pass occurrence.

        \vspace{-12pt}
        \subsubsection{Comparison of $V_{\text{frame}}$ Based on Pass Outcome.}
\begin{figure}[H]
                \centering
                \begin{minipage}{0.66\linewidth}
                    \begin{minipage}{0.48\linewidth}
                        \centering
                        \includegraphics[width=\linewidth]{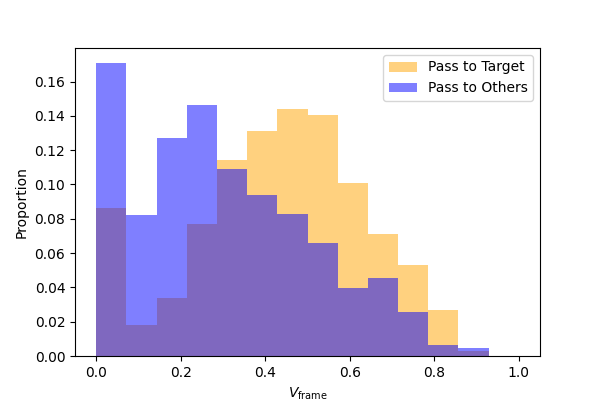}
                        \textbf{(a)} Distribution of $V_\text{frame}$: Group~1
                    \end{minipage}
                    \hfill
                    \begin{minipage}{0.48\linewidth}
                        \centering
                        \includegraphics[width=\linewidth]{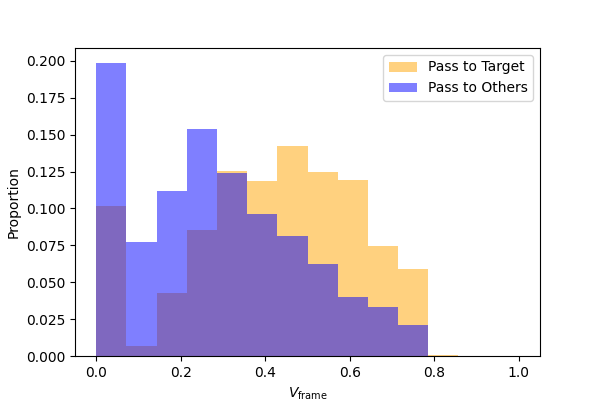}
                        \textbf{(b)} Distribution of $V_\text{frame}$: Group~2
                    \end{minipage}
                \end{minipage}

                \centering
                \begin{minipage}{0.31\linewidth}
                    \centering
                    \includegraphics[width=\linewidth]{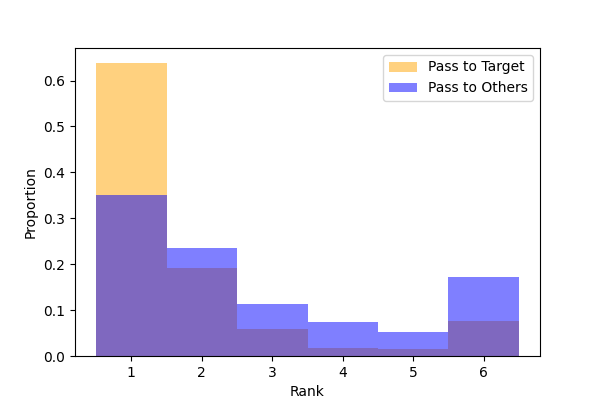}
                    \textbf{(c)} Rank in $V_\text{frame}$: Group~1
                \end{minipage}
                \hfill
                \begin{minipage}{0.31\linewidth}
                    \centering
                    \includegraphics[width=\linewidth]{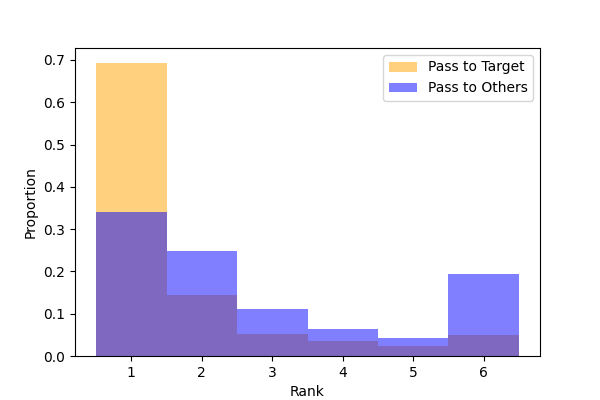}
                    \textbf{(d)} Rank in $V_\text{frame}$: Group~2
                \end{minipage}
                \hfill
                \begin{minipage}{0.31\linewidth}
                    \centering
                    \includegraphics[width=\linewidth]{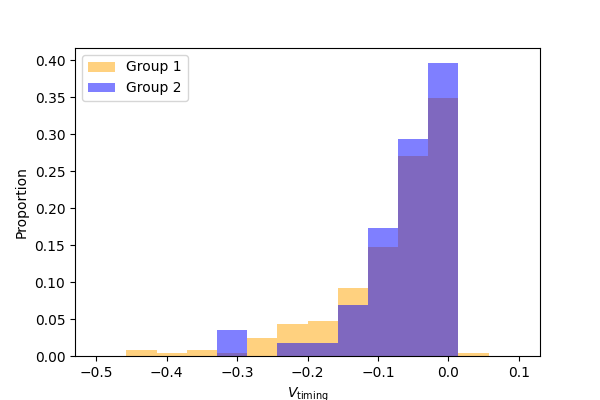}
                    \textbf{(e)} Distribution of $V_\text{timing}$: Group~1 vs. Group~2
                \end{minipage}
                \caption{Comparison of spatial and temporal evaluation metrics across skill groups and pass outcomes.  
                Distributions of $V_{\text{frame}}$ values (a, b) and team-relative ranks (c, d) in $V_{\text{frame}}$ for detected players when the pass was actually directed to them (orange) versus to another player (blue).  
                Distribution of $V_\text{timing}$ scores (e) across skill groups, where Group~1 (intermediate-to-advanced) and Group~2 (novice-to-intermediate).  
                Higher $V_{\text{frame}}$ values, higher ranks, and smaller deviations in $V_\text{timing}$ (closer to zero) support the validity of the proposed metrics.} 
                \label{fig:distribution}
            \end{figure}

            Based on the target prediction results, we compared the $V_{\text{frame}}$ values of the detected player between cases where the pass was directed to the detected player (``target'') and cases where it was directed to another player (``others'') (Fig.~\ref{fig:distribution} (a) and (b). The distributions exhibited clear separation, as confirmed by the Kolmogorov–Smirnov (KS) test. For all data, the KS D-value was 0.3147 with $p=0$, indicating a substantial difference between the two groups. When analyzed by skill group (Group 1 and Group 2; the specific criteria for group classification are described in a later section), the KS D-value was 0.3159 ($p=0$) for Group 1 and 0.3120 ($p=0$) for Group 2, both reflecting significant distributional differences.

            The D-value in the KS test represents the maximum difference between the cumulative distribution functions (CDFs) of the two groups; for example, a D-value of 0.1 indicates that the two cumulative distributions differ by at most 10\% at any point \cite{gibbons2014nonparametric}. The D-values observed in this study (all above 0.31) therefore indicate a pronounced separation between the $V_{\text{frame}}$ distributions for the two pass outcome groups. These results suggest that $V_{\text{frame}}$ tends to be higher when the detected player is the actual pass recipient, and this trend is consistent across different skill groups.

        \vspace{-12pt}
        \subsubsection{Relative Ranking Within the Offensive Team.}
            We further computed the $V_{\text{frame}}$ values for all non-possessing offensive players within each frame and ranked the detected player accordingly. This allows us to evaluate the relative spatial advantage of the detected player within the team (Fig.~\ref{fig:distribution} (c) and (d)).
            The Mann–Whitney U test indicated a statistically significant difference between cases where the detected player was the actual pass recipient and when they were not ($p=0$), and the effect size, as measured by Cliff’s Delta, was $-0.339$, which corresponds to a medium magnitude according to the interpretation criteria in \cite{meissel2024using}.

            When analyzing the data by skill group, the trend remained consistent. For Group 1, Cliff’s Delta was $-0.329$ (small magnitude), while for Group 2 it was $-0.382$ (medium magnitude), with both comparisons yielding $p=0$.
            These results indicate that, across all groups, detected players tended to rank higher in $V_{\text{frame}}$ when they were the actual pass recipient, although the effect was slightly more pronounced in Group 2.

        \vspace{-12pt}
        \subsubsection{Comparison Across Skill Groups.}
            To examine whether the $V_\text{frame}$ values are robust across different skill levels, we stratified the data into two groups:
            \begin{description}
                \item[Group~1:] Male players with 1–3 years of experience (intermediate to advanced)
                \item[Group~2:] Female players with up to 3 years and male players with less than 1 year of experience (novice to intermediate)
            \end{description}
            Figure~\ref{fig:distribution} (a) and (b) show similar distributional patterns of $V_\text{frame}$ values across the two groups, and Fig.~\ref{fig:distribution} (c) and (d) further demonstrate that the relative ranking of target players remains consistent. These parallel trends suggest that the proposed metric maintains consistent behavior regardless of player experience, indicating robustness to differences in skill level.

            However, the data used in this study were collected from a single team under limited match-like conditions. The diversity of player demographics and tactical contexts (e.g., man-to-man vs. zone defense) remains limited. Future validation across different competition levels (e.g., club team, national tournaments) and varied tactical environments would provide further evidence of the generalizability and practical utility of $V_\text{frame}$.

    \vspace{-9pt}
    \subsection{Validation of timing evaluation metric $V_\text{timing}$}
    \vspace{-5pt}
    \label{ssec:v_timing_considerations}
        This section examines the distribution of $V_\text{timing}$ (a score indicating the deviation from optimal timing) scores and explores differences in the effectiveness of initiation timing between skill groups. Figure~\ref{fig:distribution} (e) illustrates the $V_\text{timing}$ distributions for two groups categorized by experience: Group~1 (intermediate to advanced players) and Group~2 (novice to intermediate players).
        The initial hypothesis anticipated that higher-skilled players in Group~1 would be more likely to select near-optimal initiation timings, thereby achieving higher $V_\text{timing}$ values. Contrary to this expectation, the results indicated that Group~2 had a distribution concentrated closer to zero, indicating a tendency toward higher $V_\text{timing}$ values.
        Several interpretations are possible for this result. One plausible explanation is that players in Group~2 exhibit limited variability in their playing style and movement patterns. This homogeneity may constrain the range of counterfactual scenarios, thereby reducing the gap between the actual initiation and the optimal timing. 
        Another possibility is that players in Group~1, who tend to attempt more advanced plays, are subject to stronger defensive pressure. As a result, they may find it more difficult to initiate movement at the optimal moment, leading to greater deviations and thus lower $V_\text{timing}$ values despite their higher skill level.
        These considerations suggest that $V_\text{timing}$ does not merely reflect individual skill but also captures the quality of context-adaptive decision-making, taking into account the available offensive options and situational constraints such as defensive intensity. Future work should aim to refine the interpretation of $V_\text{timing}$ by incorporating contextual variables, such as defender proximity and marking status, into the counterfactual evaluation framework.

\vspace{-12pt}
\section{Conclusions} 
\vspace{-5pt}
\label{sec:conclusion}
    In this study, we proposed a novel framework called VTCS (Valuing Timing by Counterfactual Scenarios) to quantitatively evaluate how the timing of a receiver’s movement initiation affects spatial advantage in Ultimate Frisbee. VTCS constructs a series of temporally counterfactual scenarios that alter only the initiation timing of the target player and evaluates each scenario using a spatial control-based metric to assess the effectiveness of the actual decision.

    
    These findings collectively enable the quantitative assessment of movement initiation timing: an unlabelled yet tactically crucial movement that has been difficult to evaluate. While developed for Ultimate, the VTCS framework can be applied to other invasion sports where tracking data is available, thus the applications to other invasion sports as well as various competitive levels in Ultimate will be future work. 

\vspace{-8pt}
\ifarxiv
\section*{Acknowledgments}
\vspace{-8pt}
This study was financially supported by JSPS KAKENHI 23H03282, JP24K23889.
\vspace{-5pt}
\fi

\bibliographystyle{splncs04}
\bibliography{reference}

\newpage
\input{Appendix}

\end{document}

%% file: Appendix.tex
\renewcommand{\thesection}{\Alph{section}}
\appendix

\section*{Appendix}

\section{Ultimate Frisbee Rules}
\label{app:ultimate}
    Ultimate Frisbee (commonly referred to as ``Ultimate'') is an invasion-type team sport played with a flying disc. Each team consists of seven players on the field. The game is played on a rectangular field with end zones at both ends (typically 100 meters in total length, 37 meters in width, and 20-meter end zones). The offensive team aims to score one point by successfully completing a series of passes that culminate in a teammate catching the disc in the opposing end zone.

    A player holding the disc (the thrower) must establish a pivot foot and is not allowed to move from that spot while in possession of the disc until the disc is released. There is also a time limit on holding the disc, referred to as ``stalling''. A defensive player within three meters of the thrower (the marker) initiates a stall count by audibly counting ``one, two, ...'' up to ten seconds. If the disc is not thrown before the count reaches ten, a turnover is called due to a ``stall out''. Stalling functions not only as a temporal constraint but also as a defensive tactic. By controlling their positioning and stance, the marker can restrict the thrower's viable angles and limit available passing lanes. This tactical constraint reduces the offensive team’s passing options and narrows their strategic choices.

    Turnovers also occur due to dropped discs, out-of-bounds throws, or interceptions, resulting in an immediate change of possession. When a point is scored, the scoring team earns one point and begins the next point on defense by performing a ``pull''—a throw that initiates play.

    Ultimate is fundamentally non-contact and is governed by a unique ethos known as the ``Spirit of the Game'', which emphasizes self-officiating and fair play. This principle contributes to Ultimate’s distinctive combination of strategic depth and ethical sportsmanship.

\section{Related work}
\label{app:related}
    \subsection{Space evaluation methods}
        Since off-ball actions (in this case, disc reciever actions) are seldom logged as discrete events, space evaluation metrics using all players' locations are important. Early work relied on rule-based heuristics:  basketball research incorporated player profiles when rating off‑ball movements in passing sequences \cite{wu2022obtracker} and similar rule‑driven ``dangerosity'' scores have been proposed for soccer \cite{link2016real}, but a variety of off‑ball options makes general evaluation difficult.
        
        A parallel line of research models space mathematically. Dominant-region approaches use Voronoi diagrams based on minimum arrival time \cite{taki1996development,fujimura2005geometric}, later refined with player‑specific kinematics \cite{brefeld2019probabilistic,martens2021space} and distance‑weighted fields \cite{narizuka2021space}. The probabilistic Off‑Ball Scoring Opportunity (OBSO) framework \cite{OBSO} inspired indices that quantify how attackers pull defenders apart.OBSO has since been adapted to soccer (attackers \cite{teranishi2022evaluation,yeung2024strategic}, defenders \cite{umemoto2023evaluation}), basketball \cite{kono2024mathematical}, and Ultimate \cite{iwashita2024space}. Purely data‑driven variants also exist, such as a neural‑network estimator for badminton doubles \cite{ding2023estimation}.
    
    \subsection{Counterfactual analysis}
        From \cite{fujii2025machine}, machine‑learning approaches investigate ``what‑if'' questions by conditioning generative models on altered game states. In basketball, recurrent networks and diffusion models have been conditioned on teammate and opponent positions to synthesise alternative player trajectories, clarifying how specific role changes shape future movement patterns \cite{Yeh19,fujii2024decentralized,chen2023professional}. In soccer, several frameworks have been proposed to address counterfactual scenarios: the Player Action Value Estimation framework \cite{dick2021rating}, which quantifies the value of player actions by comparing predicted possession values between actual and counterfactual scenarios generated by replacing a player’s trajectory; the Shooting Payoff Computation framework \cite{yeung2024strategic}, which combines deep learning with game-theoretic reasoning to compare hypothetical shot and pass decisions; and TacticAI \cite{wang2024tacticai}, which advises corner‑kick positioning by generating counterfactual setups. 
        However, these data‑driven methods inherit the biases of their training samples and capture correlation rather than true causality when the data are sparse or unbalanced.
        
        Causal inference techniques seek stronger guarantees by estimating treatment effects from observational data. Propensity‑score matching has been applied to questions such as timeout effectiveness in basketball \cite{gibbs2020causal}, crossing pressure in soccer \cite{wu2021contextual}, and pitch selection in baseball \cite{nakahara2023pitching}. These methods can yield more valid counterfactual estimates but demand large, representative datasets and careful modeling of confounders.
        
        A third strand employs explicit mathematical models that embed sport-specific rules and physical constraints. Voronoi-based dominant-region approaches and their probabilistic successors, such as OBSO model \cite{OBSO}, provide closed‑form tools for simulating positional adjustments. Building on OBSO, Umemoto and Fujii generated counterfactual defensive alignments to identify optimal soccer defender positions \cite{umemoto2023evaluation}. Because the governing equations are predefined, such models remain interpretable and data‑efficient, offering a principled complement to purely statistical or learning‑based counterfactual frameworks.

\subsection{Evaluation of movement initiation timing}
    Early evidence on the evaluation of movement initiation comes from tightly controlled (and simulated) 1‑on‑1 basketball experiments. A series of biomechanical studies showed that defenders output earlier movement initiation when better preparatory states based on the ground reaction forces before an attacker cuts \cite{fujii2013unweighted,Fujii14,fujii2015preparatory,Fujii15,Fujii15b}.  Although these works isolate timing effects with laboratory precision, they involve only two actors and omit the spatiotemporal complexity of full team play.
    
    Field-based studies remain sport and situation specific. In soccer, dive initiation during penalty kicks has been timestamped by expert video annotation and, more recently, markerless pose estimation \cite{reddy2024identifying}. Beyond goalkeeping, \cite{herold2022off} detected the instant an attacker’s run created decisive separation, highlighting off‑ball timing as a neglected performance factor. Comparable efforts in basketball are just emerging: OBTracker mines NBA tracking data to label and score the timing of off‑ball cuts and screens, providing one of the first data‑driven baselines for temporal analysis in professional settings \cite{wu2022obtracker}. Ultimate Frisbee has virtually no timing studies, with wearable‑sensor work focusing on biomechanics rather than tactical optimisation \cite{slaughter2020tracking}, and existing state‑transition models relying on discrete pass events (e.g., \cite{lam2021state}, see also Introduction Section).
    
    Counterfactual timing frameworks are even scarcer. G‑computation has been used to estimate the effect of taking versus swinging at a 3‑0 pitch in baseball \cite{vock2018estimating}, yet the analysis ignores spatial interactions. Another basketball work employed counterfactual recurrent networks to gauge the benefit of an extra basketball pass \cite{fujii2024estimating}; however, that causal approach requires large, balanced training sets and still treats timing indirectly via sequence modelling. By contrast, our study integrates full‑field tracking with a rule‑based motion model to compute spatially explicit counterfactuals, delivering the first objective estimate of optimal movement‑initiation timing in Ultimate.

\section{Structure of the CSV file in UltimateTrack dataset}
\label{app:csv}
    See Table \ref{tab:variables}.
    \begin{table}[ht]
    \centering
    \caption{Structure of each row in the dataset CSV file}
    \label{tab:variables}
    \begin{tabular}{ll}
        \toprule
        \textbf{Variable} & \textbf{Description} \\
        \midrule
        \texttt{frame} & Frame index \\
        \texttt{id} & Object ID (1--15) \\
        \texttt{class} & Object type (\texttt{offense}, \texttt{defense}, \texttt{disc}) \\
        \texttt{x}, \texttt{y} & Position on the field [m] \\
        \texttt{vx}, \texttt{vy} & Velocity [m/s] \\
        \texttt{ax}, \texttt{ay} & Acceleration [m/s\textsuperscript{2}] \\
        \texttt{closest} & ID of the closest opposing player \\
        \texttt{holder} & Boolean indicating whether the object holds the disc \\
        \bottomrule
    \end{tabular}
    \end{table}

\section{Preprocessing}
\label{app:preprocess}
    The dataset was processed through the following preprocessing steps.
    First, the four corners of the field were manually annotated in the footage to define a coordinate system in which the offensive direction is consistently from left to right. Then, the position of each player was manually recorded frame by frame. The disc holder was annotated for each possession, and the disc position was inferred accordingly. Specifically, the disc was assumed to be at the same position as the holder during possession, and its position during passes was linearly interpolated between the previous and next known points.
    
    The image coordinates were transformed into field coordinates in meters based on the annotated reference points. The resulting position data was smoothed to estimate velocity and acceleration. Each object was then classified as offense, defense, or disc, and for each offensive player, the closest defensive player was identified and paired one-to-one.
    
    These preprocessing steps produced a high-quality tracking dataset with consistent position, velocity, and acceleration data, ready for direct use in evaluating counterfactual scenarios and conducting quantitative spatial analysis.

\section{Detection of Initiation of Movement}
\label{app:detection}
    In this study, the receiver's initiation of movement is defined as the onset of rapid acceleration with the intention of receiving a pass. We detect this event automatically using rule-based criteria. The frame where this initiation is detected becomes the starting point of a movement sequence, which is defined as the entire span of motion from that frame until the player either receives a pass or ceases the attempt. The extracted sequence is then verified via manual video inspection to eliminate false positives.

    \vspace{-12pt}
    \subsection{Definition of Initiation of Movement}
    A player is deemed to initiate movement if they satisfy all of the following conditions (Condition 1), and the corresponding frame is treated as the beginning of the movement sequence.

    \vspace{8pt}
    \noindent\hspace{2em}\textbf{Condition 1 (Criteria for detecting initiation of movement):}
    \vspace{-4pt}
    \begin{description}
      \setlength{\itemindent}{0.5em}
      \item[(i)] Acceleration magnitude is at least 4~m/s².
      \item[(ii)] The player has not held the disc for the past 2 seconds (30 frames).
      \item[(iii)] The angle between the velocity and acceleration vectors is 90° or less.
    \end{description}
    (i) A high acceleration magnitude typically reflects a decision to change movement direction.
    (ii) However, players also tend to accelerate after throwing the disc from a stationary state; to exclude such cases, we add the condition that the player has not held the disc for the preceding 2 seconds.
    (iii) Since large acceleration can also result from deceleration, we include a directional condition to ensure consistency.

    \vspace{-12pt}
    \subsection{Expansion of the Movement Sequence}
        A movement sequence is defined as the segment of motion that begins with the initiation of movement and continues until the player either receives the disc or abandons the cut. To identify the full sequence, we extend the segment forward and backward by including additional frames based on the following criteria.

        \vspace{8pt}
        \noindent\hspace{2em}\textbf{Condition 2 (Forward extension of the movement sequence):}
        \vspace{-4pt}
        \begin{description}
          \setlength{\itemindent}{0.5em}
          \item[(i)] The player was part of the movement sequence in the previous frame.
          \item[(ii)] The player is not holding the disc.
          \item[(iii)] Velocity magnitude is at least 3~m/s.
          \item[(iv)] Directional change in velocity from the previous frame is 20° or less.
          \item[(v)] Deviation from the average velocity direction in the current sequence is 90° or less.
        \end{description}

        \noindent\hspace{2em}\textbf{Condition 3 (Backward extension of the movement sequence):}
        \vspace{-4pt}
        \begin{description}
          \setlength{\itemindent}{0.5em}
          \item[(i)] The player is included in the movement sequence in the next frame.
          \item[(ii)] Velocity magnitude is at least 0.05~m/s.
          \item[(iii)] Velocity magnitude in the next frame has not decreased by more than 0.05~m/s.
        \end{description}
        Condition 2 is applied iteratively to propagate the movement sequence forward, capturing frames where the player maintains sufficient speed and directional consistency. Condition 3 is used to incorporate frames before the initiation frame, in cases where acceleration may have started earlier.

    \vspace{-12pt}
    \subsection{Exclusion Criteria}
        Among the extracted movement sequences, those satisfying the following condition are excluded from evaluation, as they are unlikely to represent attempts to receive a pass.
        
        \vspace{8pt}
        \noindent\hspace{2em}\begin{minipage}[t]{\dimexpr\linewidth-2em}
            \textbf{Condition 4 (Exclusion criteria):}\\
            \vspace{-4pt}
            At the end of the movement sequence, one or more of the following holds:
        \end{minipage}
        \begin{description}
          \setlength{\itemindent}{0.5em}
          \item[(i)] Two or more offensive players are within a 5-meter radius at the end of the movement sequence.
          \item[(ii)] Two or more offensive players are located within a 90° cone in the forward direction.
        \end{description}
        These conditions are designed to filter out movements toward crowded areas or those lacking a clear spatial objective. Additionally, all detected sequences are reviewed in video form, and those without a clear intent to receive a pass are also excluded.

\section{Definition of Space Evaluation Metric}
\label{app:space}
    \subsection{PPCF}
        PPCF defines the probability density that player $i$ controls location $\vec{r}$ at time $t$ within time $T$ as:
        \begin{equation}
            \frac{d\text{PPCF}_i}{dT}(t, \vec{r}, T) = \left(1 - \sum_{k} \text{PPCF}_k(t, \vec{r}, T)\right) f_i(t, \vec{r}, T) \lambda_i
        \end{equation}
        where $f_i(t, \vec{r}, T)$ is the probability that player $i$ can reach location $\vec{r}$ within time $T$, and $\lambda_i$ is a control ability constant (set to 4.3 for both offense and defense in this study). The final control probability field $\mathbf{PPCF}_i(t, \vec{r})$ is obtained by numerically integrating the above expression over $T \in [0, \infty)$.

    \subsection{UPPCF}
        In adapting PPCF to ultimate, we incorporate domain-specific characteristics. In ultimate, the player holding the disc is not allowed to move, and defenders positioned within 3 meters—typically engaged in stalling—also tend to remain stationary. Iwashita et al. (2024) proposed excluding these players from pitch control calculations, and we follow this approach to better reflect actual player dynamics. Additionally, we introduce a direction-dependent reaction time, assuming that reaction latency depends on the alignment between a player’s movement and the disc or their marking target. The reaction angle and reaction time are defined as follows:
        \begin{equation}
            \theta_{\text{reaction}} =
            \begin{cases}
                \theta_{d-\mathbf{v}} & \text{(offense)} \\
                \min(\theta_{d-\mathbf{v}}, \theta_{d-\mathbf{o}}) & \text{(defense)}
            \end{cases}, \quad
            \text{RT} = 0.1 + \frac{\theta_{\text{reaction}}}{\pi}
        \end{equation}
        Here, $\theta_{d-\mathbf{v}}$ is the angle between the disc direction and the player's velocity vector, and $\theta_{d-\mathbf{o}}$ is the angle to the marked offensive player.

    \subsection{Distance Weight}
    \label{app:distance_weight}
\begin{figure}
        \centering
        \includegraphics[width=0.5\linewidth]{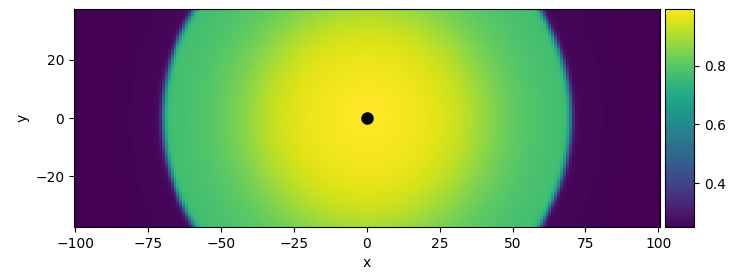}
        \caption{Spatial distribution of the distance-based weight $w_d$. The weight decreases as the distance from the disc (black dot) increases, reflecting the decreasing likelihood of successful long passes in ultimate frisbee. Warmer colors indicate higher weights.}
        \label{fig:distance_weight}
    \end{figure}

    See Fig.~\ref{fig:distance_weight}.

\section{Reachable Area}
\label{app:reachable_area}
\begin{figure}
        \centering
        \includegraphics[width=0.8\linewidth]{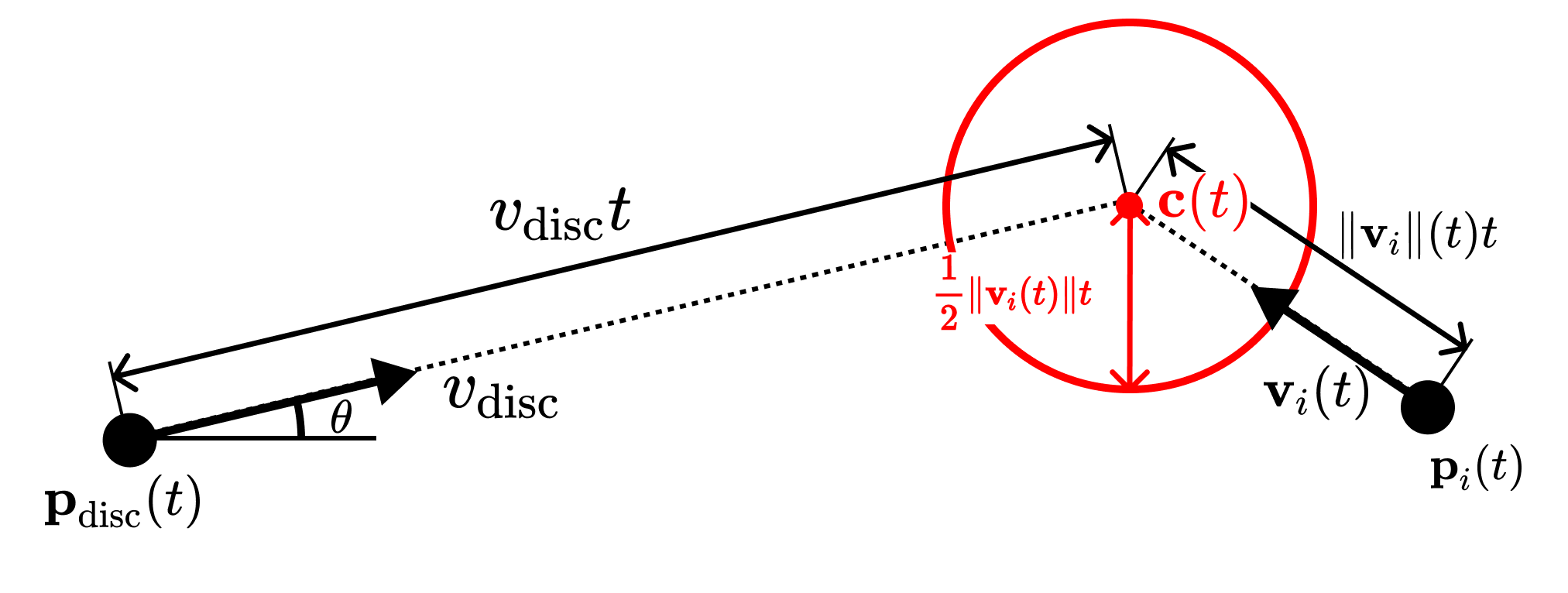}
        \caption{Geometric illustration of the predicted reachable area $\Omega(t)$, defined as a circle centered at the predicted reception point $\mathbf{c}(t)$, with a radius proportional to the player’s velocity. The disc and player trajectories are used to solve for the expected interception point.}
        \label{fig:reachable_area}
    \end{figure}

    See Fig.~\ref{fig:reachable_area}.

\vspace{-9pt}
\section{Application}
\vspace{-3pt}
\label{app:application}
    VTCS quantitatively evaluates how the timing of movement initiation influences spatial advantage by comparing actual plays with temporally counterfactual scenarios. In this section, we demonstrate its applicability through a specific example in which the initiation timing of a receiver is analyzed, and a more optimal alternative is identified.
    
\begin{figure}[htbp]
        \centering
        \begin{minipage}[b]{0.48\linewidth}
            \centering
            \includegraphics[width=\linewidth]{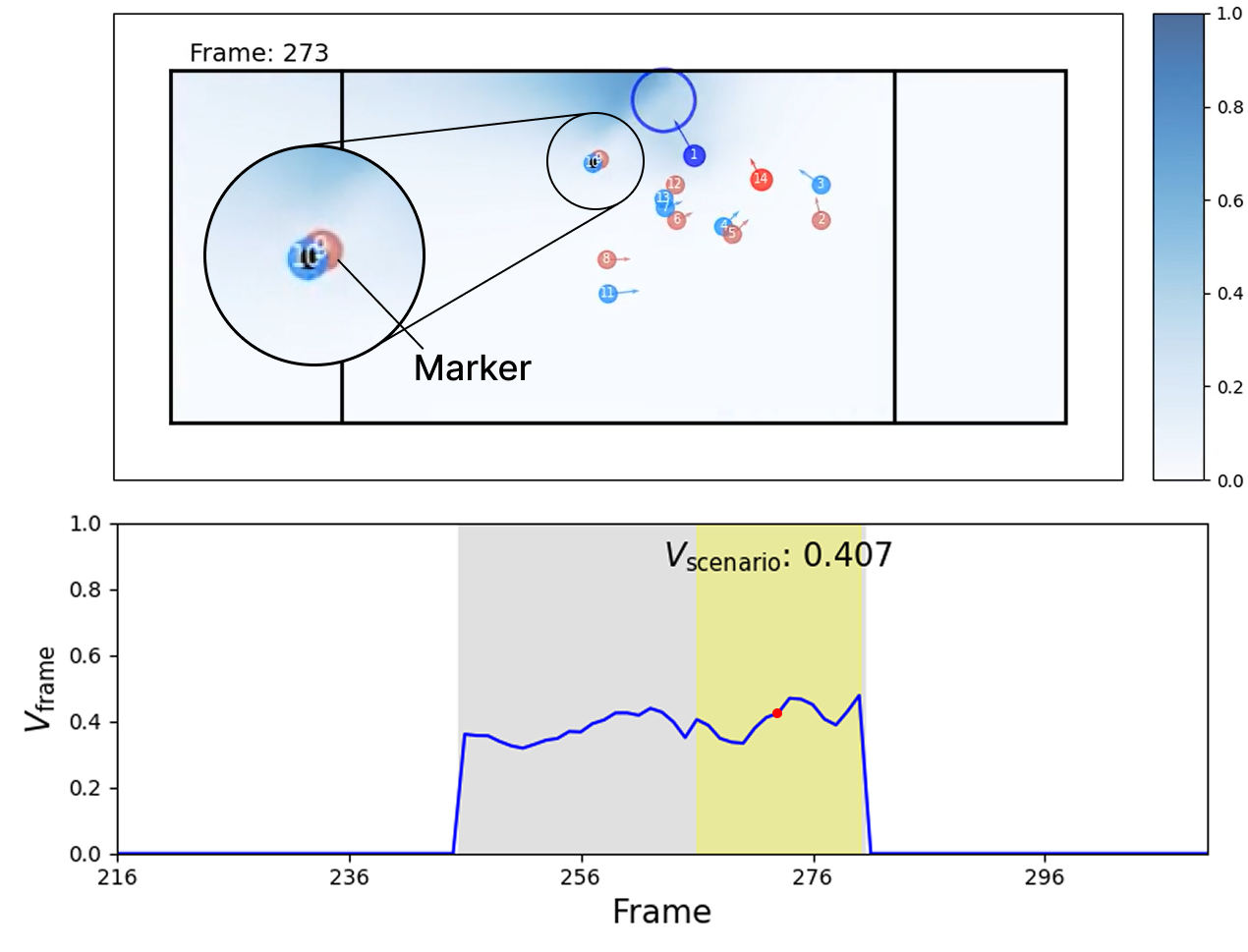}
            \textbf{(a) Actual play}\\
            The receiver has already initiated movement, but the marker is closely guarding the thrower, severely limiting the passing lane. As a result, the spatial control within the reachable area is low, and the receiver is unable to secure sufficient open space. The scenario evaluation value is $V_\text{scenario}=0.407$.
        \end{minipage}
        \hfill
        \begin{minipage}[b]{0.48\linewidth}
            \centering
            \includegraphics[width=\linewidth]{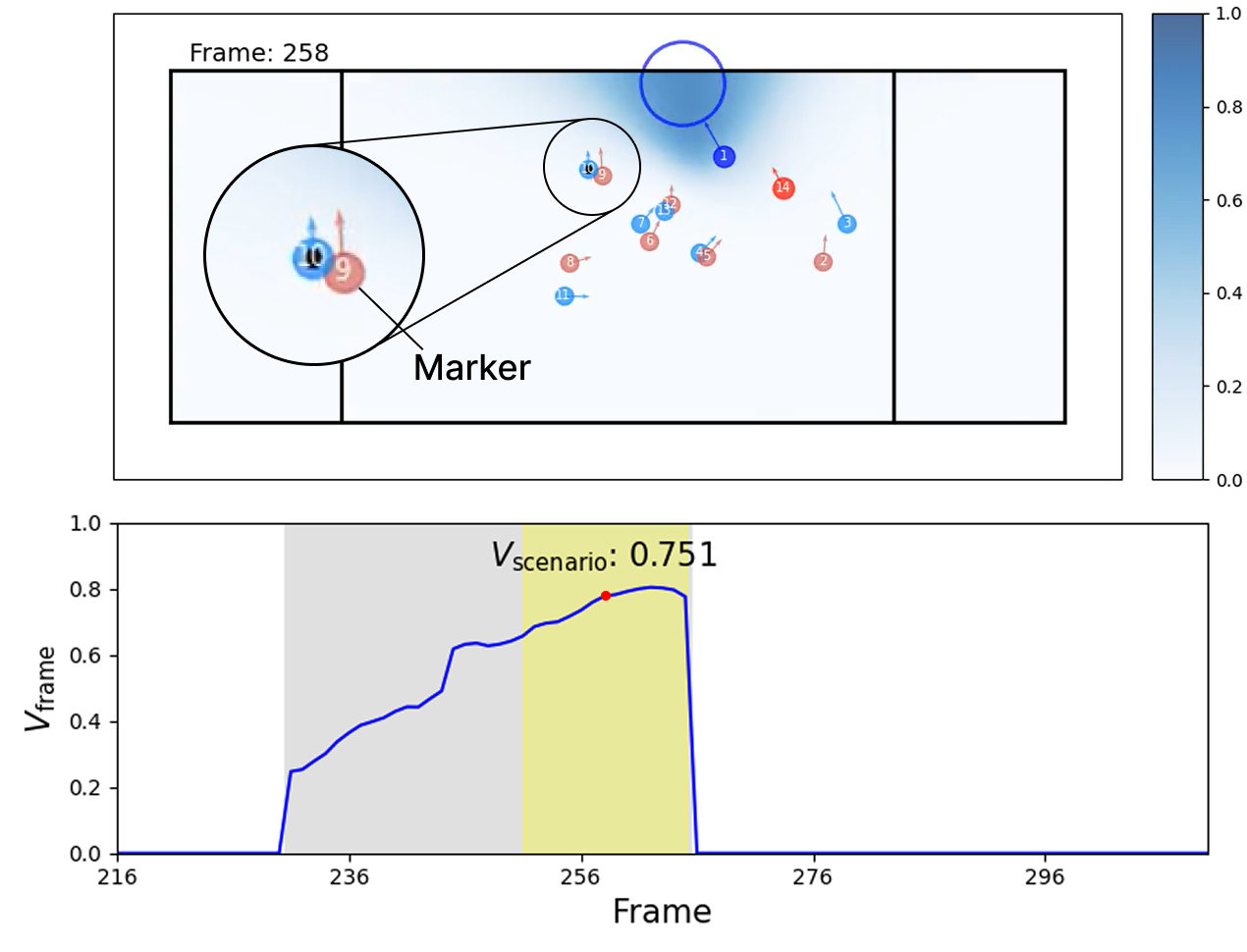}
            \textbf{(b) Optimal scenario}\\
            In the counterfactual scenario where the receiver initiates movement 15 frames earlier, they can reach open space before the marker obstructs the passing lane, thereby achieving a higher spatial advantage. The scenario evaluation value improves to $V_\text{scenario}=0.751$.
        \end{minipage}
        \caption{Comparison between the actual play (left) and the counterfactual scenario with 15-frame earlier initiation (right). The top panels show the player positions and spatial control at the respective moments; the blue circles represent the receiver’s reachable area. The bottom panels present the time series of spatial evaluation values $V_{\text{frame}}$, where the gray region indicates the evaluation window and the yellow region shows the averaging window used to compute $V_{\text{scenario}}$. The red dot marks the frame shown in the corresponding top panel. The comparison suggests that earlier initiation leads to a higher spatial advantage ($0.407 \rightarrow 0.751$).}
        \label{fig:vtcs_example}
    \end{figure}

    Figure~\ref{fig:vtcs_example} (a) illustrates the actual play. At the moment the receiver initiated movement, the marker (a defensive player near the thrower) was already approaching the thrower and limiting the passing lane, preventing the receiver from gaining sufficient open space. The scenario evaluation value at this point was $V_\text{scenario}=0.407$.

    In contrast, Fig.~\ref{fig:vtcs_example} (b) shows a counterfactual scenario in which the receiver initiated movement 15 frames (i.e., 1 second) earlier. In this scenario, the receiver was able to cut into the open space before the marker could effectively restrict the passing lane, resulting in a significantly improved spatial advantage. The scenario evaluation value in this case increased to $V_\text{scenario}=0.751$, demonstrating a clear improvement over the actual play.

    These results highlight how VTCS enables the visualization of lost spatial opportunities caused by suboptimal initiation timing. Moreover, VTCS can function not only as a diagnostic tool that provides concrete suggestions--such as how much earlier (or later) the movement should have been initiated. This insight is valuable for players and coaches seeking to improve timing decisions through data-informed feedback.

